\documentclass[journal]{IEEEtran}
\usepackage{cite}
\usepackage{times}
\usepackage{graphicx}
\usepackage{latexsym}
\usepackage{amsmath}
\usepackage{adjustbox}
\usepackage{array}
\usepackage{pgfplots}
\usepackage{amssymb}
\usepackage[vlined,linesnumbered,ruled,resetcount]{algorithm2e}

\pgfplotsset{compat=1.14}

\DeclareMathOperator*{\mean}{mean}
\DeclareMathOperator*{\argmax}{arg\,max}

\newcolumntype{R}[2]{%
    >{\adjustbox{angle=#1,lap=\width-(#2)}\bgroup}%
    l%
    <{\egroup}%
}
\newcommand*\rot{\multicolumn{1}{R{50}{1em}}}

\newcommand{\specialcell}[2][c]{%
  \begin{tabular}[#1]{@{}c@{}}#2\end{tabular}}
 
\newcommand{\specialcellSTD}[2][l]{%
  \begin{tabular}[#1]{@{}l@{}}#2\end{tabular}}

\makeatletter
\newcommand{\removelatexerror}{\let\@latex@error\@gobble}
\makeatother

\begin{document}

\title{Spatio-temporal Tubelet Feature Aggregation and Object Linking for Video Object Detection}

\author{Daniel~Cores,
        V\'ictor~M.~Brea,
        Manuel~Mucientes
\thanks{The authors are with Centro Singular de Investigación en Tecnoloxías Intelixentes (CiTIUS), Universidade de Santiago de Compostela, Santiago de Compostela, 15782, Spain (e-mail: daniel.cores@usc.es; victor.brea@usc.es; manuel.mucientes@usc.es).}}

%

\maketitle

\begin{abstract}
This paper addresses the problem of how to exploit spatio-temporal information available in videos to improve the object detection precision. We propose a two stage object detector called FANet based on short-term spatio-temporal feature aggregation to give a first detection set, and long-term object linking to refine these detections. Firstly, we generate a set of short tubelet proposals containing the object in $N$ consecutive frames. Then, we aggregate RoI pooled deep features through the tubelet using a temporal pooling operator that summarizes the information with a fixed size output independent of the number of input frames. On top of that, we define a double head implementation that we feed with spatio-temporal aggregated information for spatio-temporal object classification, and with spatial information extracted from the current frame for object localization and spatial classification. Furthermore, we also specialize each head branch architecture to better perform in each task taking into account the input data. Finally, a long-term linking method builds long tubes using the previously calculated short tubelets to overcome detection errors. We have evaluated our model in the widely used ImageNet VID dataset achieving a 80.9\% mAP. Also, in the challenging small object detection dataset USC-GRAD-STDdb, our proposal outperforms the single frame baseline by 5.4\% mAP.
\end{abstract}


%
\IEEEpeerreviewmaketitle

\section{Introduction}
\IEEEPARstart{O}{bject} detection has been one of the most active research topics in computer vision for the past years.  However, the use of temporal information in videos to boost the detection precision is still an open problem. Although a traditional object detection framework can be executed at frame level, it does not exploit temporal information available on videos that can be crucial to address challenges such as motion blur, occlusions or changes in objects appearance at certain frames. 

In general, object detection frameworks implement two main tasks: bounding box regression and object classification. The bounding box regression is responsible for the spatial object localization. Therefore, in video processing, the current frame at each time instant provides the most valuable information to perform this task. In contrast, we hypothesize that extracting spatio-temporal information from the appearance of the object in previous frames can improve significantly the classification task accuracy. This raises the issue of linking and aggregating spatio-temporal features throughout time.

State-of-the-art object detection frameworks are based on two major approaches: one stage and two stage architectures. In one stage methods \cite{liu2016ssd}, \cite{redmon2016yolo}, the network head has to process a dense set of candidate object locations, with a high imbalance between objects of interest and background sample. Two stage frameworks  \cite{girshick2015fast}, \cite{girshick2014rich}, \cite{ren2015faster} address this issue adding an object proposal method as the first stage that filters out most of the candidates associated with background. Then, the network head refines the proposal set. We develop a two stage spatio-temporal architecture in which the proposals generated by the first stage are also used to propagate information from previous frames.

The proposed framework is an extension of the Faster R-CNN model to use both temporal and spatial information to enhance the network classification accuracy. As the original Faster R-CNN, our deep convolutional network can be trained end-to-end without any precomputed proposals. The main novelties of our proposal are:

\begin{itemize}

    \item A tubelet proposal method that works with Feature Pyramid Network (FPN) models dealing with multiple Region Proposal Networks (RPN) and extracting RoI features at different pyramid levels. As far as we know, this is the first spatio-temporal framework with a multiple-level backbone such as FPN.
    
    \item A temporal pooling method able to summarize information from the previous $N$ frames producing a feature map with the same size as if the network was working with a single frame. Thus, it works with a fully connected network head with the same number of parameters and a constant execution time independently of $N$.
    
    \item A spatio-temporal double head. One branch takes spatial information from the current frame to perform the bounding box regression and the spatial classification, and another branch complements the classification output using information from the previous $N$ frames.
    
    \item A long-term optimization method creates long tubes associating object instances throughout the video, rescoring all detections belonging to the same tube. This method reuses short-term information to improve the long tubes creation process, overcoming network errors at certain frames such as missing detections that could otherwise break the tubes.
    
\end{itemize}

    

A preliminary version of this spatio-temporal framework was described in \cite{ecai2020}. The main differences in the CNN come from the new double head architecture capable of better combining spatio-temporal information with the spatial information extracted from the current frame. Moreover, we extend the experimentation evaluating new training and testing strategies. All these changes lead to an improvement of 2.7\% mAP with respect to \cite{ecai2020} in the widely used ImageNet VID dataset. We also add a small object detection dataset (USC-GRAD-STDdb) \cite{bosquet2018stdnet} to assess the performance of our network by providing new scenarios that are not covered by Imagenet VID. We obtain promising results with our general purpose framework on a very specific problem such as the small object detection\cite{tong2020recent}.

\section{Related work}
The object detection problem was first defined in the single image domain following two main approaches: two stage and one stage architectures. Lately, spatio-temporal frameworks were proposed based on these methods but taking into account the temporal information available in videos to improve the detection precision.

Two stage architectures were first popularized by R-CNN \cite{girshick2014rich}. R-CNN needs to apply a Convolutional Neural Network (CNN) on each region of a precalculated proposal set, resulting in a very slow approach. This issue is addressed in Fast R-CNN \cite{girshick2015fast}, adding an RoI pooling layer on top of the CNN. Instead of executing the CNN over each proposal, it takes the whole image as input, generating a global deep feature map. Then, the RoI pooling generates a per proposal feature map extracting the corresponding features. This way, it improves dramatically both the training and test times by sharing all the CNN backbone calculations.

All these methods rely on an external region proposal method. The Faster R-CNN framework \cite{ren2015faster} defines an RPN to generate the proposal set in a fully convolutional fashion reusing the backbone calculations. This makes it possible to perform an end-to-end training of the whole system without any precomputed information. The Feature Pyramid Network (FPN) \cite{lin2017feature} proposes a change in the definition of the CNN backbone extracting feature maps at different depth levels instead of taking just the deepest one. Therefore, the RPN and the network head have to calculate object proposals and the final detections at different feature map levels. FPN implements a top-down pathway and lateral connections to combine low-resolution, semantically strong features with high-resolution, semantically weak features to be able to work at different levels without loosing semantic meaning.

The R-FCN object detector \cite{dai2016rfcn} re-implements the network header avoiding the fully connected layers used by previous work. Instead, it follows a fully convolutional approach changing the RoI pooling by a position sensitive RoI pooling. Many other works propose new header implementations based on the original Faster R-CNN framework such as the Cascade R-CNN \cite{cai2018cascade}. This method defines a multi-stage head that refines the object proposal set iteratively. Wu \textit{et al.} \cite{wu2020rethinking} studied how the head architecture affects the bounding box regression and the classification tasks proposing a double head approach to take advantage of the benefits of each architecture in each task.

One stage detectors such as SSD \cite{liu2016ssd} and YOLO \cite{redmon2016yolo} do not refine a previously calculated proposal set. Instead, they generate a dense grid of bounding boxes and directly calculate the final detection set. Recent works \cite{lin2017focal} try to overcome the object/background imbalance problem  modifying the cost function to prevent easy examples from having a huge impact on the network training process.

Since the introduction of the ImageNet object detection from videos (VID) challenge, the video object detection problem has drawn the attention of the research community. Even so, how temporal information available in videos should be used to improve the detection performance remains an open problem. In fact, the same issue also remains unsolved in the action recognition field. For instance, Mhalla \textit{et al.} \cite{mhalla2019spatio} propose to generate interlaced images containing information from several video frames to feed the object detector.

Two-stream networks such as \cite{carreira2017quo}, \cite{feichtenhofer2016spatiotemporalaction}, \cite{feichtenhofer2016convolutionaltwostream} or \cite{simonyan2014twostream} have become the standard approach in action recognition. One of the branches processes video frames, while the other one takes precomputed dense optical flow frames as input. Adeli \textit{et al.} \cite{adeli2019component} define a three stream architecture that also takes into account the background context. Although action recognition is a related problem, the benefits of adding optical flow information to spatio-temporal object detectors might not be so straightforward. To be able to distinguish some action classes such as \emph{"sitting down"} and \emph{"getting up"}, motion information given by optical flow might be crucial. This is not so evident in object detection. For instance, authors in \cite{kalogeiton2017joint} conclude that traditional methods for mixing movement and appearance information work for actions but not for objects using the same network architecture.

Still, optical flow information has proven successful in \cite{zhu2017flow} for video object detection. This work proposes to aggregate deep spatial features throughout time to improve the per-frame feature maps. To do so, it utilizes the movement information given by the optical flow to find the correspondences between the current features and the nearby feature maps. We propose a novel method to find these correspondences working with two stage frameworks by linking object proposals throughout time.

Several approaches have been proposed to link object detections throughout neighbouring frames making up short object tubelets. Object tracking has been used to link detections generated by a frame level object detector in \cite{kang2016object}. T-CNN \cite{kang2017t} works on two single frame detectors and also applies tracking to link these detections over time. Kang \textit{et al.} \cite{kang2017object} define a Tubelet Proposal Network (TPN) with two main components. First, it propagates static proposals at frame level across time. Then, the second network component calculates the bounding box displacement in each frame to build the tubelet proposal. Although this second component works with pooled features extracted using the same bounding box across time, the network can handle moving objects due to the generally large receptive field of CNNs.

Alternatively, the idea of \emph{anchor box} defined in single frame object detectors can be extended to the spatio-temporal domain. The ACtion Tubelet detector (ACT-detector) \cite{kalogeiton2017action} utilizes \emph{anchor cuboids} to initialize the action tubelets. Tang \textit{et al.} \cite{tang2019object} propose a Cuboid Proposal Network (CPN) as the first step for short object tubelet detection. Unlike these previous methods, in our implementation, each \emph{anchor box} in the \emph{anchor cuboid} is regressed independently by the corresponding RPN with information from the corresponding frame, allowing us to reuse part of the computations from previous iterations while preserving the proposals linked throughout consecutive frames.

The aim of the described methods is to link objects in the short-term. Therefore, they only take into account the nearby frames wasting long-term information. To overcome this, Feichtenhofer \textit{et al.} \cite{feichtenhofer2017detect} use tracking information as input to an object linking algorithm to build long tubes, and aggregate detection scores throughout the tube. To do that, tracking and object detection are performed and learned simultaneously using a multi-task objective training. Munjal \textit{et al.} \cite{munjal2020joint} extend this idea learning a per object unique re-identification feature vector embedding that improves the long-term tracking performance.

Tang \textit{et al.} \cite{tang2019object} also perform a long-term object linking concatenating small tubelets. First, they calculate short temporally overlapping tubelets so one single frame could have detections associated with more than one tubelet. Then, tubelets are joined by analyzing the overlap in the shared frames.

All these methods only consider the final detections set to perform the long-term object linking. In this work, we propose to include short-term information calculated by the RPN to overcome missed detections, being able to build larger tubes. In doing so, we only use object proposals and detections given by the network without any external tracking method to aid the object linking process.

\section{FANet Architecture}

\begin{figure*}
\centerline{\includegraphics[width=\textwidth]{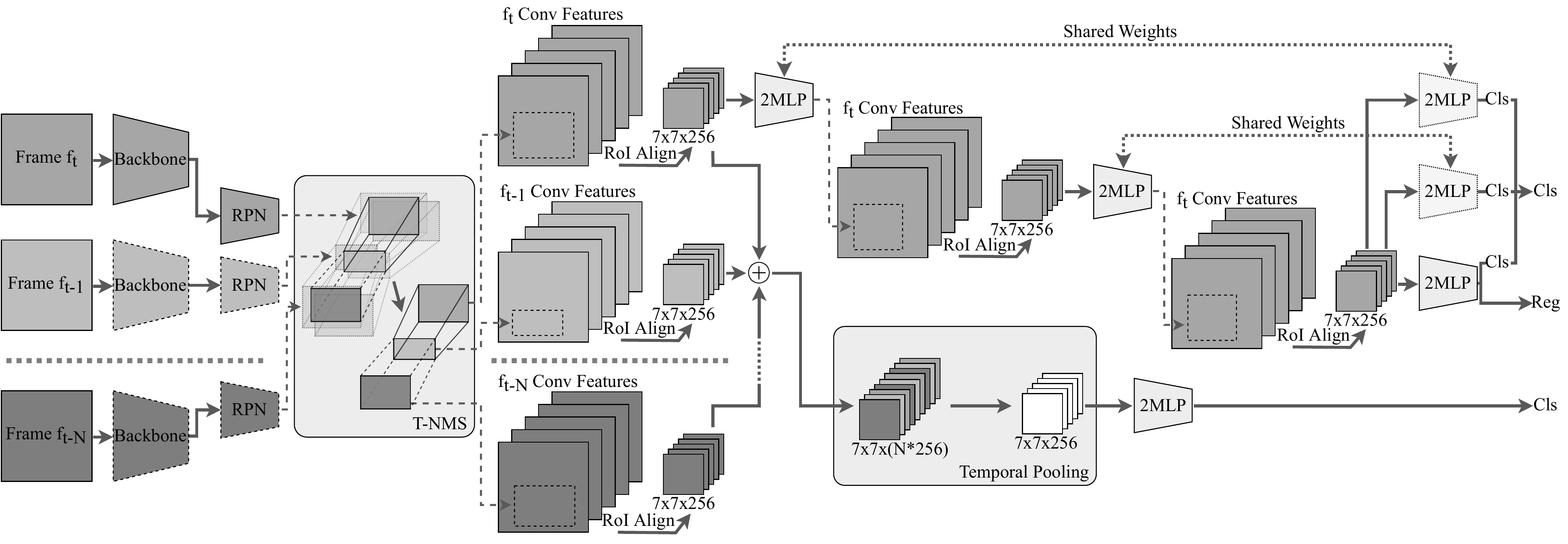}}
\caption{
FANet architecture with a single level backbone. Dotted backbone and RPN boxes represent components that are reused without new computations. Blocks labeled as 2MLP in the network head implement a multilayer perceptron with two fully connected layers.}

\label{fig:architecture}
\end{figure*}

The proposed framework (Fig.\ref{fig:architecture}) generalizes a single frame two stage object detector to take advantage of working with a sequence of $N$ frames $f_{t-N-1}, ..., f_{t-1}, f_t$ to improve the accuracy in each frame $f_t$. Even though we build our system on Faster R-CNN \cite{ren2015faster} with a Feature Pyramid Network (FPN) backbone \cite{lin2017feature} to illustrate the network architecture, the same ideas could be applied to other models. Indeed, since the spatio-temporal tubelet proposal is a core concept in our architecture, the multi-scale level approach imposes higher complexity than single scale models due to the multiple Region Proposal Networks (RPN). These object tubelets link proposals throughout time allowing the network to combine object information from multiple temporal instants.

We initialize object tubelets as a sequence of $N$ \emph{anchor boxes}, one per frame, in the same position, with the same area and aspect ratio. Then, the RPN modifies each \emph{anchor box} independently in the corresponding frame using features calculated by the corresponding backbone. This, together with the fact that both the network backbone and the RPN share the convolutional weights among all input frames, allows us to reuse the backbone and RPN computations to reduce the overhead associated with our spatio-temporal approach with respect to the single frame method. This way, feature maps and object proposals associated with frame $f_{t-1}$ at time $t$ become feature maps and proposals associated with frame $f_{t-2}$ at time $t+1$. As a result, this process outputs a set of tubelet proposals that must be filtered to remove the spatially redundant ones. To do that, we add a Tubelet Non-Maximum Suppression (T-NMS) \cite{tang2019object} algorithm on top of the RPNs. The tubelet generation is further addressed in Sec.~\ref{seq:tubelet_proposals}.

Having the tubelet proposals and the backbone features, the RoI Align method \cite{he2017mask} extracts a per frame and per tubelet feature map of fixed size centered on the object. Therefore, features from these feature maps can be directly associated by position and propagated throughout previous frames. Fig.~\ref{fig:architecture} shows the simplest case in which we have a single level backbone rather than the more complex FPN. After the RoI Align, we concatenate all feature maps associated with proposals belonging to the same tubelet. Then, we shuffled the channels, so channels in the same position in the original feature maps are consecutive in the concatenated one (Fig.\ref{fig:architecture}). The resulting feature map has a joined dimension $N$ times the original RoI Align size, making it dependent on the number of input frames. The Temporal Pooling method reduces this dimension to a fixed size independently of $N$. Sec.~\ref{seq:temporal_pooling} describes the joining and pooling processes in detail.

We introduce a double head approach to process both the current frame and the spatio-temporal information (Sec.~\ref{seq:double-head}). The spatial branch is fed by the RoI Align output in the current frame, while the spatio-temporal branch takes the Temporal Pooling output. Since the spatial branch is mainly intended to localize the object in current frame, we follow a multi-stage object architecture \cite{cai2018cascade} in which each stage output is the input to the next. This way, it refines object proposals gradually to maximize the overlap with the actual object. The complete proposed network can be trained end-to-end without heavily engineered proposals.

Lastly, per frame object detections are linked making up long tubes. Short-term object tubelets provide helpful information about whether or not two detections in different neighbouring frames are the same object. The proposed long-term object linking algorithm takes this information as input to grow the final tubes. Then, classification score is propagated throughout each object tube (Sec.~\ref{sec:tubes}).

\subsection{Tubelet proposals}
\label{seq:tubelet_proposals}
In general, the starting point for most object detectors is a set of predefined \emph{anchor boxes}. Then, they adjust these \emph{anchor boxes} to better fit the object and assign an object category removing those classified as background. Similarly, tubelet proposals are initialized as a sequence of $N$ \emph{anchor boxes}, generating \emph{anchor cuboids} \cite{kalogeiton2017action} in our spatio-temporal framework. Each \emph{anchor box} in the \emph{anchor cuboid} has exactly the same size, aspect ratio, and is in the same position for all input frames. Therefore, the number of \emph{anchor cuboids} remains the same as the number of \emph{anchor boxes} in the single frame approach, distributing $k$ \emph{anchor boxes} at each sliding position $W \times H$, generating $W \times H \times k$ \emph{anchor cuboids} in total.

The FPN (Feature Pyramid Network) backbone \cite{lin2017feature} distributes the \emph{anchor boxes} among the RPNs by area. In our implementation, every single \emph{anchor box} $b^i$ in the \emph{anchor cuboid} sequence $(b^{t-N-1}, ...,b^{t-1}, b^t)$ is regressed independently in its corresponding frame at the appropriate level. This means that, the RPN can reuse the bounding box regression calculations for $b^i \in (b^{t-N-1}, ..., b^{t-1})$, and only needs to calculate $b^t$ at the current instant. Adapting this strategy to single level models just implies that \emph{anchor boxes} belonging to all \emph{anchor cuboids} are being mapped to the same level.

Regressing every \emph{anchor cuboid} leads to spatially redundant tubelets in the proposal set. In two stage single frame frameworks, this problem is generally solved executing the Non Maximum Suppression (NMS) method over the proposal set. In our case, doing a per frame NMS might break the tubelets, removing some of their bounding boxes. Instead, we apply an extension of the NMS algorithm called Tubelet Non-Maximum Suppression (T-NMS) based on \cite{tang2019object}, but with different metrics to calculate the tubelet score $ts(\tau_i)$ and the overlap between two tubelets $\tau_i$ and $\tau_j$. This method aims to remove spatially redundant tubelets instead of single boxes.

The score of a given tubelet $\tau_i$ is calculated as:

\begin{equation}
    ts(\tau_i) = mean(bs_i^{t-N-1}, bs^{t-1}_i, ..., bs^t_i).
    \label{eq:tubelet_score}
\end{equation}
being $bs^t$ the score associated with the proposal $b$ at frame $t$.

The overlap between a pair of tubelets $\tau_i$ and $\tau_j$ is defined as: 

\begin{equation}
    overlap(\tau_i, \tau_j) = \mean_{k=t-N-1,..., t} IoU(b_i^k, b_j^k).
\end{equation}

Unlike the original FPN strategy which performs a per level NMS, our T-NMS implementation removes globally the spatially redundant proposals, taking as input the whole set of tubelets. The resultant subset $\mathcal{T}$ represents the final collection of proposals.

\subsection{Temporal Pooling}
\label{seq:temporal_pooling}
\begin{figure*}
\centerline{\includegraphics[width=1\textwidth]{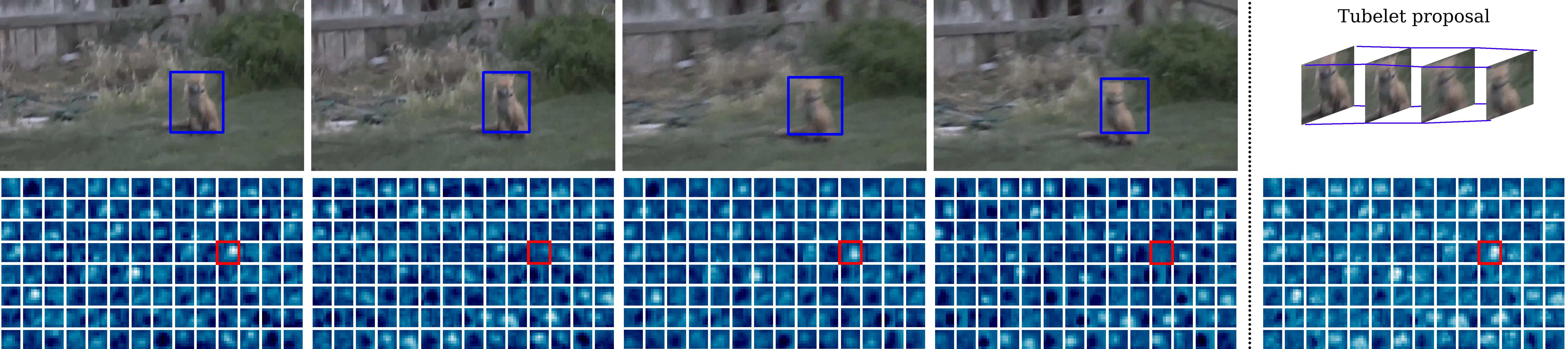}}
\caption{Temporal pooling example with a number of input frames $N=4$. From left to right, there are the four input frames with an object proposal (the blue bounding box) in each one, belonging all of them to the same tubelet. Under each input frame, a subset of channels from the RoI Align output are represented. On the right, we show the tubelet proposal linking all box proposals. At the bottom right, we also show the aggregated feature map calculated by our Temporal Pooling method. We highlight one channel (surrounded in red) as an example of how the highest activations (lighter colors) in each frame contribute to the aggregated feature map.}
\label{fig:features_propoagation}
\end{figure*}

In Faster R-CNN based object detectors, an RoI feature pooling method takes the proposal set to produce a  per proposal fixed size feature map. Working with FPN, object proposals are distributed among the pyramid levels according to their size to perform the RoI pooling over the corresponding feature map. We use the RoI Align\cite{he2017mask} method to perform the feature pooling operation taking each bounding box $b^j$ belonging to each tubelet $\tau_i = (b^{t-N-1}_i, ..., b^{t-1}_i, b^t_i)$ to extract features from the corresponding pyramid level in the frame $f_j$. Performing this mapping process independently in each frame rather than per tubelet allows to map each box $b^j$ within the same tubelet $\tau_i$ to a different pyramid level making our system robust against scale variations in the tubelet sequence. As a result, we have a fixed size feature map (in our case of $7 \times 7 \times 256$, Fig.~\ref{fig:architecture}) associated with each box in the tubelet.

On top of that, we introduce an operator called \emph{Temporal Pooling} that summarizes the whole tubelet information generating a feature map with just the same size as the RoI Align output for a single frame, independently of the number of input frames $N$. This method requires $N$ to be small enough to allow the RPN to adapt the corresponding \emph{anchor box} in the \emph{anchor cuboid} sequence to fit the object in each frame. Working with a large $N$, the object movement might exceed the RPN receptive field, making impossible to adjust the same \emph{anchor box} in every frame to the target object \footnote{As Section~\ref{sec:ablation_studies} shows, the network achieves the best result with $N=6$.}. Since, all RoI Align outputs associated with the same tubelet have a fixed size and are centered in occurrences of the same object over time, we can associate features in the same position of the RoI pooled feature map in every frame.

Although errors in the RPN bounding box regression might cause feature mismatches, our method is robust against small variations since we use pooled features. These pooled features are a broad representation of features extracted from the backbone, so slight changes on the proposal bounding box do not have a significant effect in the RoI Align output.

Firstly, to perform the temporal aggregation, our method concatenates the $N$ input feature maps with size $W \times H \times C$ into one feature map with size $W \times H \times N \cdot C$ (Fig.~\ref{fig:architecture}). Then, we reorder the resulting feature map channels, so that channels at the same position in all the input feature maps are concatenated consecutively (see the input to the Temporal Pooling operator in Fig.~\ref{fig:architecture}). Finally, the Temporal Pooling applies the following transformation to get each element of the resultant feature map $X$:

\begin{equation}
    \label{eq:max_pooling}
    x_{ijk} = \max_{t=0,...,N-1}(y_{ij(N(k-1)+t)})
\end{equation}
being $x_{ijk}$ an element in the position $i \times j$ in channel $k$ of the output feature map, and $y_{ij(N(k-1)+t)}$ an element in the position $i\times j$ in channel $(N(k-1)+t)$ from the concatenated feature map. As Fig.~\ref{fig:features_propoagation} shows, this propagates the highest activation values throughout the tubelet.

\subsection{Spatio-temporal double head}
\label{seq:double-head}
The network head follows a double head approach based on the idea that aggregated spatio-temporal information is valuable to improve object classification while spatial information extracted from the current frame is crucial for bounding box regression. Consequently, we design each head branch to perform better in their respective task taking into account the input data.

The spatial head (Fig.~\ref{fig:architecture} at the upper right) takes as input the RoI Align output in the current frame $f_t$ to calculate a spatial object classification and a class agnostic bounding box regression. Since the main goal of this branch is the object localization, we implement a cascade of object detectors \cite{cai2018cascade} to iteratively refine the object proposal set until the final bounding box regression is done.

Following the Cascade R-CNN training strategy, we perform a proposal resampling after each stage applying an increasing IoU threshold to assign each proposal to a ground truth object. This way, the requirements to consider one proposal as positive example to train the corresponding stage are harder as we advance in the cascade. In general, an IoU too high might assign all proposals to background vanishing the positive examples. Nevertheless, as each stage takes as input the refined proposal set by the previous stage, we can increase the IoU achieving more accurate boxes in each stage. On test time, we use the average of the classification scores calculated by every stage detector over the final proposal set \cite{cai2018cascade}.

The spatio-temporal head (Fig.~\ref{fig:architecture} at the bottom right) classifies the object based on the output of the Temporal Pooling operator. Therefore, this branch takes into account the appearance of the object in the previous $N$ frames.

As a result, this strategy produces a bounding box regression and two object classification vectors, one calculated using features in the current frame and another calculated using aggregated features through the previous $N$ frames. The final classification is calculated as follows \cite{wu2020rethinking}:

\begin{equation}
    \label{eq:double_head}
    p = p_{tmp} + p_{spt}(1 - p_{tmp})
\end{equation}
where $p_{spt}$ and $p_{tmp}$ are the score vectors from the spatial and temporal heads, respectively. As a result, we are considering spatial classification in the current frame and spatio-temporal classification at the same level. That makes the current frame $f_t$ to have a greater influence in the final classification result than any previous frame $f_i$ in $f_{t-N-1}, ..., f_{t-1}$.

\subsection{Long-term object linking}
\label{sec:tubes}

\begin{figure*}
\centerline{\includegraphics[width=1\textwidth]{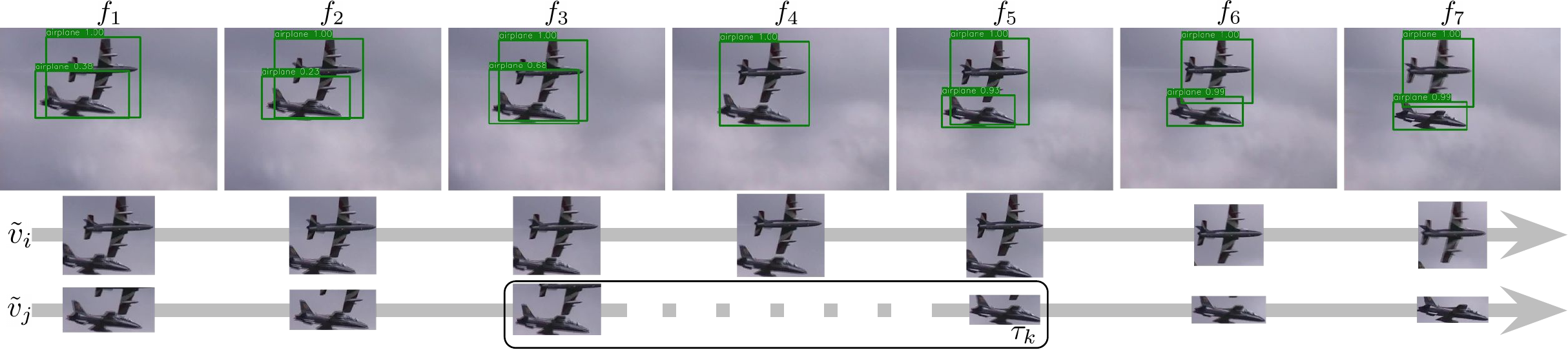}}
\caption{Long-term object linking. Green boxes are actual network detections. The network does not detect the two objects in $f_4$ breaking $\tilde{v}_j$ in two fragments. The last detection of the first fragment and the first detection of the second fragment belong to the same RPN tubelet ($\tau_k$).}
\label{fig:tubes}
\end{figure*}

We propose a two step long-term object linking algorithm that takes network detections and produces long object tubes. Linking network detections over time to identify single action/object instances has become a standard approach in both action recognition \cite{gkioxari2015finding}, \cite{saha2016deep}, and recently in object detection \cite{feichtenhofer2017detect}, \cite{kang2017t}, \cite{tang2019object}. These methods try to join single boxes or small tubelets into larger tubes in order to propagate the classification score throughout the video.

In the first stage of our method, the frame to frame linking problem is defined as an optimization problem that maximizes the accumulated linking score in each tubelet. Network errors such as false negatives or misclassified detections might brake large tubes since it is not possible to find a detection to link in some frames. The second step of the long term linking method utilizes the short-term tubelet information to overcome some of these problems allowing the algorithm to produce larger tubes.

In this implementation, each detection $d_t^i = \{x_t^i, y_t^i, w_t^i, h_t^i, p_t^i\}$ in the set $D_t$ indexed by $i$ in the frame $t$, is composed of a box centered at $(x_t^i, y_t^i)$ with width $w_t^i$ and height $h_t^i$, and an associated classification confidence $p_t^i$ for the object class. A lower threshold $\beta$ is applied over the detection set $D_t$ before the object linking algorithm to prevent that low confidence detections adversely affect the tube creation. The linking score $ls(d^i, d^j)$ between two detections $d^i$ and $d^j$ at different frames $t$ and $t'$ is defined as:

\begin{equation}
    ls(d^i_t, d^j_{t'}) = p^i_t + p^j_{t'} + IoU(d^i_t, d^j_{t'}).
\end{equation}

Then, the optimal tube $\hat{v}$ can be found by solving the following optimization problem applying the Viterbi algorithm per object category:
\begin{equation}
    \hat{v} = \argmax_{\mathcal{V}} \sum_{t=2}^{T} ls(D_{t-1}, D_t)
    \label{eq:viterbi}
\end{equation}
where $\mathcal{V}$ is the set of all possible tubes.

Algorithm~\ref{alg:viterbi} describes how we create long-term object tubes in detail. This method solves Equation~\ref{eq:viterbi} (Algorithm~\ref{alg:viterbi}:\ref{alg:viterbi_eq}) iteratively to find all tubes ending in frame $i$, starting with tubes ending in the last frame $i=T$. In each iteration, all detections belonging to the best tube $\hat{v}$ are removed from the candidate set $\mathcal{D}$ (Algorithm~\ref{alg:viterbi}:\ref{alg:viterbi_remove}) and the new tube is added to the result set (Algorithm~\ref{alg:viterbi}:\ref{alg:viterbi_add}). When there are no more candidate detections in the current frame (Algorithm~\ref{alg:viterbi}:\ref{alg:viterbi_fdone}), the same process is applied to find all tubes ending in the previous frame (Algorithm~\ref{alg:viterbi}:\ref{alg:viterbi_for}). The process ends when there are no more tubes ending in the second frame of the video. This is due to the fact that we directly create a tube of length $|\hat{v}|=1$ for each detection in the first frame that does not belong to any tube computed in previous iterations.

\begin{figure}[!t]
 \removelatexerror
\begin{algorithm}[H]
    \SetKwInOut{Input}{Input}
    \SetKwInOut{Output}{Output}
    \SetKw{KwOR}{or}
    \SetKw{KwAND}{and}
    \SetKw{KwNOT}{not}
    \SetKw{KwIN}{in}
    \SetKw{KwDONE}{done}
    \SetKwFunction{FVit}{Viterbi}

    \Input{Per frame detection set $\mathcal{D} = \{D_t = (d_t^1, ..., d_t^{n_t})\}_{t=1}^T$}
    \Input{All possible tubes: $\mathcal{V}$}
    \Output{Object tubes $\hat{\mathcal{V}}$}
    
    $\hat{\mathcal{V}} \gets \emptyset$
    
    \For{$i$ \KwIN $T, ..., 2$}
    {\label{alg:viterbi_for}
        \While{$D_i \neq \emptyset$}
        {\label{alg:viterbi_fdone}
            
            $\hat{v} \gets \argmax_{\mathcal{V}} \sum_{t=2}^{i} ls(D_{t-1}, D_{t})$\label{alg:viterbi_eq}
            
            $\mathcal{D} \gets \mathcal{D} \setminus \hat{v}$\label{alg:viterbi_remove}
            
            $\hat{\mathcal{V}} \gets \hat{\mathcal{V}} \cup \hat{v}$\label{alg:viterbi_add}
        }
    }
    \caption{Long-term tubes creation}\label{alg:viterbi}
\end{algorithm}
\end{figure}

Previous works defined action/objects tubes as sequences of consecutive detection boxes, without interruptions. In fact, the output of the method described above follows this definition of object tube. Nevertheless, object occlusions or even network errors such as false negatives or misclassified examples might artificially break large tubes into smaller chunks.

Information given by RPN tubelets can be used to link small tubes to provide larger ones, making up the second step of the long-term object linking. In Fig.~\ref{fig:tubes}, both the last detection of the first fragment and the first detection of the second fragment of $\tilde{v}_j$ belong to the same RPN tubelet. Having this information into account, our algorithm links the two fragments making one larger tube.

\begin{figure}[!t]
 \removelatexerror
\begin{algorithm}[H]
    \SetKwInOut{Input}{Input}
    \SetKwInOut{Output}{Output}
    \SetKw{KwOR}{or}
    \SetKw{KwAND}{and}
    \SetKw{KwNOT}{not}
    \SetKw{KwIN}{in}
    \SetKw{KwDONE}{done}
    \SetKwFunction{FVit}{Viterbi}

    \Input{Per frame detection set $\mathcal{D} = \{D_t = (d_t^1, ..., d_t^{n_t})\}_{t=1}^T$}
    \Input{Tubelet set $\mathcal{T} = \{\tau_{i} = (b_i^1, ..., b_i^N)\}_{i=1}^\theta$}
    \Input{Object tubes $\hat{\mathcal{V}} = \{\hat{v}_{i} = (d^{i,1}, ..., d^{i,m_i}\}_{i=1}^\delta$}
    \Output{Joined object tubes $\tilde{\mathcal{V}}$}
    
    $\tilde{\mathcal{V}} \gets \hat{\mathcal{V}}$
    
    \For{$\hat{v}_i$ \KwIN $\hat{\mathcal{V}}$}
    {
        \For{$\hat{v}_j$ \KwIN $\hat{\mathcal{V}}$}
        {
            $ts_{max}=0$
            
            \For{$\tau_l$ \KwIN $\mathcal{T}$}{
                \If{$\exists b_l^k \in \tau_l \mid \gamma(b_l^k, d^{i, m_i})$ \KwAND $ \exists b_l^{k'} \in \tau_l \mid \gamma(b_l^{k'}, d^{j, 1})$ \KwAND $time(d^{i, m_i}) > time(d^{j, 1})$}{
                \label{alg:linking_tubelet}
                    \If{$ts(\tau_l) > ts_{max}$}{
                        \label{alg:linking:max_score:start}
                        $ts_{max} = ts(\tau_l)$\label{alg:linking:max_score:end}
                    }
                }
            }
            $\mathcal{C}_{ij} = ts_{max}$
        }
    }
    $\mathcal{H} \gets \textit{Hungarian}(\mathcal{C})$ \label{alg:linking:hungarian}
    
    \For{$h_i$ \KwIN $\mathcal{H}$}
    {
        $\tilde{\mathcal{V}} \gets \tilde{\mathcal{V}} \setminus \hat{v}_{h_i}$ \label{alg:linking:hungarian_remove}
        
        $\tilde{v}_{i} \gets \tilde{v}_{i} \cup \hat{v}_{h_i}$ \label{alg:linking:hungarian_end}
    }

    \For{$\tilde{v}_{i}$ \KwIN $\tilde{\mathcal{V}}$}{
        $updateScores(\tilde{v}_{i})$\label{alg:linking:updateScores}
    }
    
    \caption{Long-term object tube linking}\label{alg:linking}
\end{algorithm}
\end{figure}

Algorithm~\ref{alg:linking} describes the tube linking method in detail. Formally, giving two tubes $\hat{v}_i=(d^{i,1}, ..., d^{i,m_i})$ and $\hat{v}_j=(d^{j,1}, ..., d^{j,m_j})$ with size $m_i$ and $m_j$ respectively, both will be joined in a tube of size $m_i+m_j$ if $d^{j,1}$ follows $d^{i,m_i}$ in time, and both detections belong to the same RPN tubelet (Algorithm \ref{alg:linking}:\ref{alg:linking_tubelet}). Thus, the tubelet allows to link both tubes as it contains detections from both of them, although the tubes do not have temporal overlap.

The detection set $D_t$ is the output of Non-Maximum Suppression (NMS) followed by a Bounding Box Voting transformation \cite{gidaris2015object} to the actual network output. This method takes the highest score detection $d$ and removes all other detections with an overlap with $d$ higher than a given threshold. Therefore, the final detection $d$ has many associated network outputs. Since each network detection came from one RPN tubelet, $d$ has also many RPN tubelets associated, one per each suppressed detection. Therefore, there can be many tubelets $\tau$ that contain the first or the last detection box of a specific tube. The function $\gamma(b_l^k, d^{i, m_i})$ returns True if the detection $d^{i, m_i}$ is associated with the box proposal $b_l^k$ in the tubelet $\tau_l$ (Algorithm \ref{alg:linking}:\ref{alg:linking_tubelet}).

For a given tube $\hat{v}_i=(d^{i,1}, ..., d^{i,m_i})$ we might have more than one candidate tube $\hat{v}_j$ to join with. We use the RPN tubelet score defined by Equation~\ref{eq:tubelet_score} to choose the best candidate to link for $\hat{v}_i$. We choose the highest score of all tubelets associated with $d^{i,m_i}$ and $d^{j,1}$ (Algorithm~\ref{alg:linking}:\ref{alg:linking:max_score:start}-\ref{alg:linking:max_score:end}). Then, a cost matrix $\mathcal{C}$ can be constructed with as many rows as ending candidate tubes and as many columns as starting tubes. Each element $c_{ij}$ is the maximum score for tubelets containing $d^{i,m_i}$ and $d^{j,1}$. This becomes an assignment problem that can be easily solved with the \emph{Hungarian Method}  (Algorithm~\ref{alg:linking}:\ref{alg:linking:hungarian}-\ref{alg:linking:hungarian_end}). For each tube assignment $(i, j)$, we remove the second tube from the output set (Algorithm~\ref{alg:linking}:\ref{alg:linking:hungarian_remove}) and build a new tube joining the two original fragments (Algorithm~\ref{alg:linking}:\ref{alg:linking:hungarian_end}). Once this process has finished, for a giving tube $\tilde{v}_i$ the score of all detections $d^{i,j} \in \tilde{v}_i$ is set to the mean of the $\alpha=10\%$ highest scores in that tube (Algorithm~\ref{alg:linking}:\ref{alg:linking:updateScores}).

\section{EXPERIMENTS}
\subsection{Datasets}
\label{sec:datasets}
We use the ImageNet VID dataset \cite{ILSVRC15} to evaluate the proposed framework and how each component contributes to the final result. This dataset contains 30 different objects categories annotated in 3,862 training and 555 validation videos. As the test subset annotations are not publicly available and the challenge evaluation server is closed, we use the Average Precision (AP) and Mean AP over the validation set as the main evaluation metrics following the standard approach established by previous works \cite{feichtenhofer2017detect,kang2016object,tang2019object,zhu2017flow}.

To train the single frame baseline we select 20 uniformly distributed frames over time from each video in the VID training set. Nevertheless, the spatio-temporal network needs $N$ consecutive input frames in each iteration instead of just one. To do that, we select 2 groups of input frames with length $N \times 15$. We have trained our spatial baseline also with this sampling strategy to assess that it does not bias the results analysis. We have not seen any significant differences in the precision values of the test for the two strategies in the single frame case.

Following the training procedure proposed by \cite{feichtenhofer2017detect}, we also use data from the ImageNet DET dataset to enhance the training set. ImageNet DET consists of 456,567 training and 20,121 validation images of 200 different categories that include the 30 object classes used in VID. We add to the training set at most 2,000 images per VID object class from ImageNet DET. This upper bound prevents the bias of the training set for large object categories in DET. In the spatio-temporal case, we transform these images into videos of $N$ repeated frames.

In addition to the ImageNet VID dataset, we also test our proposal with the Small Target Detection database (USC-GRAD-STDdb) \cite{bosquet2018stdnet} to evaluate the proposed framework in a completely different scenario. This dataset contains 115 videos with over 56,000 annotated small objects of 5 different categories. Small objects are defined in USC-GRAD-STDdb as those that fit in a bounding box with an area range between 16 ($\approx 4 \times 4$) to 256 ($\approx 16 \times 16$) pixels. This makes a huge difference between both datasets since all anotations in ImageNet VID have an area greater than 256 pixels.

\subsection{Implementation details}
The chosen backbone is a Feature Pyramid Network (FPN) based on ResNeXt-101 \cite{xie2017aggregated} pretrained on the ImageNet classification dataset. We add to both the single frame and the spatio-temporal approach a 3 stage cascade of detectors\cite{cai2018cascade} as described in Fig.~\ref{fig:architecture}. To train our spatio-temporal framework, we first train the single frame model and then we initialize the spatio-temporal network with the same weights keeping all learned layers frozen. Thus, we only have to train the spatio-temporal head if we have the equivalent single frame model already trained.

All input images are scaled so that the smallest dimension is 720 at most. If the highest dimension remains above 1,280 pixels, the image is scaled down to prevent that. In any case, the image scaling keeps the original aspect ratio.

For the spatial baseline, we use the SGD learning algorithm with a learning rate set to $2.5 \times 10^{-4}$ for the first 240K iterations, $2.5 \times 10^{-5}$ for the next 80K iterations, and $2.5 \times 10^{-6}$ for the last 40K iterations. RPN redundant proposals are suppressed by NMS with a threshold of 0.7, while the final detection set is filtered by means of NMS with an IoU threshold equal to 0.5. 

To train the spatio-temporal head we set the learning rate to $1.25 \times 10^{-3}$ for the first 180K iterations, $1.25 \times 10^{-4}$ for the next 60K iterations, and $1.25 \times 10^{-5}$ for the last 30K iterations. The network needs $N$ input images for each example: the current frame and the $N-1$ previous ones. For this reason, we replicate the first frame $N-1$ times to be able to process the first $N-1$ frames of each video.

Finally, we apply a Bounding Box Voting transformation \cite{gidaris2015object} and a confidence threshold $\beta = 0.05$ over the output detection set. This filtering prevents detections with lower scores from degrading the long-term object linking output.

\subsection{Ablation studies}
\label{sec:ablation_studies}
We conducted a series of experiments to asses how the number of input frames affects the network precision. Moreover, we also perform a collection of ablation studies in order to better understand how each network component contributes to the final result.

\begin{figure}
\begin{center}
\begin{tikzpicture}
\begin{axis}[
    width=6cm,height=4cm,
    nodes near coords={\pgfmathprintnumber[fixed zerofill, precision=1]{\pgfplotspointmeta}},
    grid=major,
    xlabel = Train/Test $N$,
    ylabel = mAP,
    ymin=75,   ymax=77.5,
    cycle list name=black white,
    title=Short-term Tubelet Length,
    xtick={2,...,5}
]

\addplot+ coordinates {
	(2,76.0)
	(3,76.4)
	(4,76.7)
	(5,76.9)
}; 

\end{axis}
\end{tikzpicture}    
\end{center}
\caption{Detection mAP when training and testing with the same tubelet lengths ($N$) on the ImageNet VID validation set without long-term information.}
\label{fig:num_frames}
\end{figure}
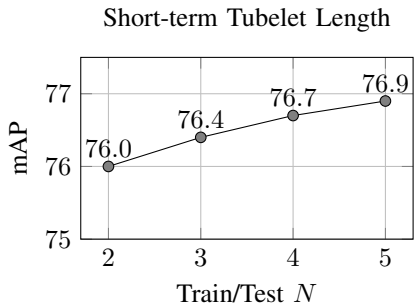

\begin{figure}
\begin{center}
\begin{tikzpicture}
\begin{axis}[
    width=8.5cm,height=4cm,
    nodes near coords={\pgfmathprintnumber[fixed zerofill, precision=1]{\pgfplotspointmeta}},
    grid=major,
    xlabel = Test $N$,
    ylabel = mAP,
    ymin=75,   ymax=77.5,
    cycle list name=black white,
    title=Short-term Tubelet Length,
    xtick={2,...,9}
]

\addplot+ coordinates {
	(2,76.0)
	(3,76.4)
	(4,76.6)
	(5,76.8)
	(6,76.9)
	(7,76.9)
	(8,76.9)
	(9,76.9)
}; 

\end{axis}
\end{tikzpicture}    
\end{center}
\caption{Detection mAP when training with $N=2$ and testing with different tubelet lengths ($N$) on the ImageNet VID validation set without long-term information.}
\label{fig:2f_nf}
\end{figure}
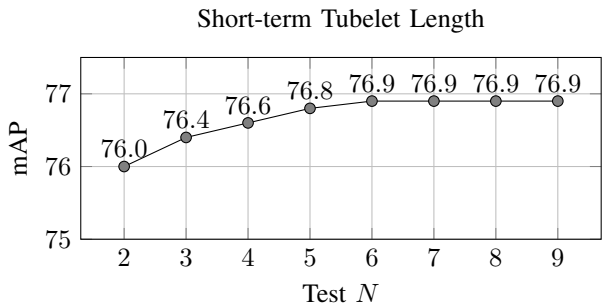

Fig.~\ref{fig:num_frames} shows how the number of input frames $N$ influences the network precision on the ImageNet VID validation set. Our current training implementation keeps all backbone computations for every input frame in GPU memory, making unfeasible to train with more than $N=5$ in a single Nvidia Tesla P40 24GB. However, since the number of parameters in our framework is independent of the number of input frames, we can tests the network with a higher $N$ independently of the training configuration. Fig.~\ref{fig:2f_nf} shows the network precision training with $N=2$ input frames and testing up to a maximum of 9 frames. By comparing Fig.~\ref{fig:num_frames} and Fig.~\ref{fig:2f_nf}, we can conclude that training and testing with a different number of input frames barely affects the network mAP, compared with training and testing with the same $N$, showing a maximum difference of 0.1\% mAP for $N=4$ and $N=5$.
 
Fig.~\ref{fig:2f_nf} shows how increasing the number of input frames up to 6 leads to an improvement in the network precision, proving that our approach exploits spatio-temporal information in the short-term. Fig.~\ref{fig:2f_nf} also shows that increasing $N$ above 6 has no impact in mAP. This might be because of the tubelet initialization based on \emph{anchor cuboids}: it expects that the same object can be associated with the same \emph{anchor box} in the same position in every input frame. If the object movement exceeds the network receptive field this assumption is not true, which is more likely for large $N$ values.

\begin{table}
    \caption{Influence of each component on the framework precision on ImageNet VID dataset.}
    \label{tab:ablation_studies}
    \centering
    \begin{tabular}{cccc|c}
        \hline
        \specialcell{Spatial\\head Cls} &  \specialcell{Spatio-temporal\\head Cls} & \specialcell{Cascade\\head} & \specialcell{Long-term\\object linking} & Mean AP \\ \hline
        \checkmark & & & & 74.4\\
        & \checkmark & & & 73.7\\
        \checkmark & \checkmark & & & 76.0\\
        \checkmark & & \checkmark & & 75.3\\
        & \checkmark & \checkmark & & 73.9\\
        \checkmark & \checkmark & \checkmark & & 76.9\\ \hline
        \checkmark & \checkmark & & \checkmark & 78.5\\
        \checkmark & \checkmark & \checkmark & \checkmark & 78.6\\ \hline
    \end{tabular}
\end{table}

\begin{table*}
    \tabcolsep=0.17cm
    \caption{VID validation set results. We use a number of input frames $N=6$ to test our framework.}
    \label{tab:resultVID}
    \centering
    \begin{tabular}{l|cccccccccccccccc}
    \hline
    ~& \rot{airplane} & \rot{antelope} & \rot{bear} & \rot{bicycle} & \rot{bird} & \rot{bus} & \rot{car} & \rot{cattle} & \rot{dog} & \rot{domestic cat} & \rot{elephant} & \rot{fox} & \rot{giant panda} & \rot{hamster} & \rot{horse} & \rot{lion} \\ \hline
    Kang \textit{et al.} \cite{kang2017object} & 84.6 & 78.1 & 72.0 & 67.2 & 68.0 & 80.1 & 54.7 & 61.2 & 61.6 & 78.9 & 71.6 & 83.2 & 78.1 & 91.5 & 66.8 & 21.6 \\ 
    Kang \textit{et al.} \cite{kang2017t} & 83.7 & 85.7 & 84.4 & 74.5 & 73.8 & 75.7 & 57.1 & 58.7 & 72.3 & 69.2 & 80.2 & 83.4 & 80.5 & 93.1 & 84.2 & 67.8 \\
    Lee \textit{et al.} \cite{lee2016multi} & 86.3 & 83.4 & 88.2 & 78.9 & 65.9 & 90.6 & 66.3 & 81.5 & 72.1 & 76.8 & 82.4 & 88.9 & 91.3 & 89.3 & 66.5 & 38.0\\ 
    Feichtenhofer \textit{et al.} \cite{feichtenhofer2017detect} & 90.2 & 82.3 & 87.9 & 70.1 & 73.2 & 87.7 & 57.0 & 80.6 & 77.3 & 82.6 & 83.0 & 97.8 & 85.8 & 96.6 & 82.1 & 66.7 \\ 
    Wang \textit{et al.} \cite{wang2018fully} & 88.7 & 88.4 & 86.9 & 71.4 & 73.0 & 78.9 & 59.3 & 78.5 & 77.8 & 90.6 & 79.1 & 96.3 & 84.8 & 98.5 & 77.4 & 75.5 \\ 
    Tang \textit{et al.} \cite{tang2019object} & 90.5 & 80.1 & 89.0 & 75.7 & 75.5 & 83.5 & 64.0 & 71.4 & 81.3 & 92.3 & 80.0 & 96.1 & 87.6 & 97.8 & 77.5 & 73.1 \\ \hline
    FPN-X101 baseline & 91.7 & 82.3 & 81.3 & 68.8 & 74.8 & 80.0 & 62.8 & 57.4 & 72.0 & 80.5 & 75.9 & 88.8 & 88.7 & 89.9 & 75.0 & 54.8 \\
    FPN-X101 Cascade baseline & 93.1 & 83.3 & 82.2 & 72.0 & 75.0 & 82.1 & 63.7 & 64.3 & 71.9 & 77.1 & 77.4 & 86.2 & 87.5 & 85.9 & 74.5 & 58.3 \\
    FANet (Short-Term) & 93.0 & 84.5 & 82.8 & 72.5 & 75.3 & 83.0 & 63.4 & 64.5 & 73.8 & 79.3 & 77.8 & 87.1 & 87.6 & 89.0 & 74.3 & 62.5 \\
    FANet (Short\&Long-Term) & 90.2 & 85.0 & 85.1 & 72.2 & 76.0 & 84.7 & 62.3 & 68.2 & 79.4 & 86.2 & 77.5 & 88.3 & 86.4 & 94.8 & 76.6 & 70.1 \\
    FANet$^\star$ (Short-Term) & 93.7 & 85.2 & 86.1 & 78.1 & 75.0 & 84.6 & 66.8 & 71.0 & 76.1 & 86.2 & 79.2 & 88.0 & 88.3 & 96.4 & 74.2 & 66.3 \\
    FANet$^\star$ (Short\&Long-Term) & 90.4 & 85.4 & 88.2 & 77.0 & 76.5 & 87.3 & 65.5 & 76.6 & 81.5 & 92.5 & 79.3 & 89.6 & 87.8 & 98.8 & 77.2 & 74.3 \\\hline \hline
    ~ & \rot{lizard}   & \rot{monkey}   & \rot{motorcycle} & \rot{rabbit}  & \rot{red panda} & \rot{sheep} & \rot{snake} & \rot{squirrel} & \rot{tiger} & \rot{train} & \rot{turtle} & \rot{watercraft} & \rot{whale} & \rot{zebra} & \multicolumn{2}{c}{Mean AP} \\ \hline
    Kang \textit{et al.} \cite{kang2017object} & 74.4 & 36.6 & 76.3 & 51.4 & 70.6 & 64.2 & 61.2 & 42.3 & 84.8 & 78.1 & 77.2 & 61.5 & 66.9 & 88.5 & \multicolumn{2}{c}{68.4} \\
    Kang \textit{et al.} \cite{kang2017t} & 80.3 & 54.8 & 80.6 & 63.7 & 85.7 & 60.5 & 72.9 & 52.7 & 89.7 & 81.3 & 73.7 & 69.5 & 33.5 & 90.2 & \multicolumn{2}{c}{73.8} \\ 
    Lee \textit{et al.} \cite{lee2016multi} & 77.1 & 57.3 & 88.8 & 78.2 & 77.7 & 40.6 & 50.3 & 44.3 & 91.8 & 78.2 & 75.1 & 81.7 & 63.1 & 85.2 & \multicolumn{2}{c}{74.5}\\
    Yang \textit{et al.} \cite{yang2016ILSVRC} & - & - & - & - & - & - & - & - & - & - & - & - & - & - & \multicolumn{2}{c}{76.2} \\
    Feichtenhofer \textit{et al.} \cite{feichtenhofer2017detect} & 83.4 & 57.6 & 86.7 & 74.2 & 91.6 & 59.7 & 76.4 & 68.4 & 92.6 & 86.1 & 84.3 & 69.7 & 66.3 & 95.2 & \multicolumn{2}{c}{79.8} \\
    Zhu \textit{et al.} \cite{zhu2017flow} & - & - & - & - & - & - & - & - & - & - & - & - & - & - & \multicolumn{2}{c}{80.1} \\
    Wang \textit{et al.} \cite{wang2018fully} & 84.8 & 55.1 & 85.8 & 76.7 & 95.3 & 76.2 & 75.7 & 59.0 & 91.5 & 81.7 & 84.2 & 69.1 & 72.9 & 94.6 & \multicolumn{2}{c}{80.3} \\
    Bertasius \textit{et al.} \cite{bertasius2018object} & - & - & - & - & - & - & - & - & - & - & - & - & - & - & \multicolumn{2}{c}{80.4} \\
    Xiao and Lee \cite{xiao2018video} & - & - & - & - & - & - & - & - & - & - & - & - & - & - & \multicolumn{2}{c}{80.5} \\
    Tang \textit{et al.} \cite{tang2019object} & 81.5 & 56.0 & 85.7 & 79.9 & 87.0 & 68.8 & 80.7 & 61.6 & 91.6 & 85.5 & 81.3 & 73.6 & 77.4 & 91.9 & \multicolumn{2}{c}{80.6}\\\hline
    FPN-X101 baseline & 78.6 & 55.2 & 85.8 & 66.7 & 68.6 & 60.1 & 59.2 & 53.8 & 89.6 & 84.2 & 77.2 & 72.0 & 75.2 & 90.5 & \multicolumn{2}{c}{74.7} \\
    FPN-X101 Cascade baseline & 78.0 & 55.6 & 84.7 & 65.8 & 81.5 & 62.1 & 64.7 & 57.3 & 89.9 & 87.1 & 76.0 & 75.0 & 76.7 & 89.3 & \multicolumn{2}{c}{75.9} \\
    FANet (Short-Term) & 78.8 & 57.3 & 85.2 & 63.1 & 82.8 & 62.3 & 68.2 & 59.5 & 90.5 & 87.2 & 76.8 & 76.2 & 76.8 & 90.4 & \multicolumn{2}{c}{76.9}\\
    FANet (Short\&Long-Term) & 80.7 & 57.5 & 88.7 & 71.2 & 89.3 & 62.9 & 69.3 & 62.4 & 90.6 & 85.0 & 77.6 & 76.9 & 72.4 & 90.6 & \multicolumn{2}{c}{78.6}\\
    FANet$^\star$ (Short-Term) & 80.4 & 58.3 & 86.2 & 71.1 & 85.1 & 59.0 & 75.7 & 60.8 & 92.0 & 87.8 & 78.9 & 76.6 & 77.0 & 90.7 & \multicolumn{2}{c}{79.2} \\
    FANet$^\star$ (Short\&Long-Term) & 82.7 & 58.2 & 89.7 & 78.2 & 91.8 & 62.0 & 76.1 & 64.3 & 91.5 & 84.6 & 80.0 & 77.0 & 72.2 & 90.7 & \multicolumn{2}{c}{80.9} \\\hline
    \end{tabular}
\end{table*}

Table~\ref{tab:ablation_studies} details how each network component contributes to the final mean average precision (mAP) testing on the ImageNet VID validation set. We use in all the experiments the bounding boxes calculated by the spatial header, choosing the classification scores from the spatial branch, the spatio-temporal branch or the combination of both (Eq.~\ref{eq:double_head}). Regarding the bounding boxes, we also compare the network precision with and without refining the object proposals using a cascade of detectors. In these experiments, using only the classification score calculated by the spatial head is different from just running the spatial network baseline. This is due to the T-NMS method that filters the proposal set in a different fashion than the conventional NMS.

Table~\ref{tab:ablation_studies} shows how the combination of the spatial and the spatio-temporal data  outperforms the execution of each of them independently. It happens with and without applying a cascade refinement, proving that they provide complementary information in both cases. The cascade bounding box refinement improves the mAP in every case. As expected, this precision gain has less influence when working only with spatio-temporal classification, as it only has an effect over the bounding box regression. In the other cases, the benefits of having this cascade also influence the spatial classification, resulting in a higher mAP improvement.

Finally, Table~\ref{tab:ablation_studies} also shows how the long-term object linking highly improves the final mAP. This is consistent with previous works that implement this kind of post-processing methods, and reveals that the RPN tubelet information can be valuable to link object instances throughout the video. Nevertheless, applying long-term object linking reduces the improvement of the cascade head. This might be due to the fact that improving the detection set calculated without cascade is more straightforward, while the version with cascade produces more accurate detections and are, therefore, harder to improve.

\subsection{Results}

\subsubsection{ImageNet VID}

We compare our framework with state-of-the-art spatio-temporal object detectors in Table~\ref{tab:resultVID}. T-CNN \cite{kang2017t} is the winner of the ILSVRC2015 with 73.8\% mAP on the validation set. This work uses two single frame object detector frameworks to calculate the initial detection set and then applies context suppression and detection propagation using optical flow. Then, it utilizes visual tracking to make up the object tubelets. Visual tracking techniques are also used in Multi-Class Multi-Object Tracking (MCMOT) \cite{lee2016multi}, the ILSVRC2016 winner \cite{yang2016ILSVRC} (they are able to boost up the system precision from 76.2\% to 81.2\% mAP with model ensembles) and in Detect to Track and Track to Detect \cite{feichtenhofer2017detect}. Our system relies on the definition of \emph{anchor cuboids} to build tubelets, instead of the more complex visual tracking algorithms, outperforming the best tracking-based method \cite{feichtenhofer2017detect} by 1.1\%.

Flow-Guided Feature Aggregation (FGFA) \cite{zhu2017flow} is a core component of the ILSVRC2017 winner, achieving 80.1\% mAP without bells-and-whistles on the validation set. Instead of building object tubelets, they find feature correspondences based on optical flow information. Finding correspondences based on optical flow is also present in the Motion-Aware Network (MANet) proposed by Wang \textit{et al.} \cite{wang2018fully}. Alternatively, Bertasius \textit{et al.} \cite{bertasius2018object} define a spatio-temporal Sampling Network (STSN) based on deformable convolutions to establish feature correspondences across the video getting 80.4\% mAP. We propose to find those correspondences by associating RoI pooled features through object tubelets, avoiding to calculate the optical flow, and getting 80.9\% mAP.

Xiao and Lee \cite{xiao2018video} propose a Recurrent Neural Network (RNN) based on a Spatial-Temporal Memory Module (STMM) that aggregates appearance information throughout time. The STMM modifies the spatial-temporal memory taking as input deep feature maps from each frame and the previous memory state, which carries information of all previous frames. This memory contains similar information to the temporal pooling output in our implementation, aggregating spatial features over time. The extra complexity of adding a RNN over the CNN backbone does not represent a clear benefit, getting slightly worse results than our method.

Tang \textit{et al.} \cite{tang2019object} implement a short tubelet detection framework to identify tubelets with temporal overlapping. Then, given two tubelets, they join those tubelets analyzing the spatial overlap between bounding boxes belonging to each tubelet in the common frame. They achieve 80.6\% mAP.

Our framework outperforms all methods in Table~\ref{tab:resultVID} achieving 80.9\% mAP with a precision gain of 5.0\% mAP compared to our single frame baseline (\emph{FPN-X101 Cascade} in Table~\ref{tab:resultVID}). We report our results with multi-scale testing ---marked with $^\star$ in Table~\ref{tab:resultVID}--- following \cite{tang2019object}, and without testing augmentation.

\subsubsection{USC-GRAD-STDdb}
\begin{table}
    \tabcolsep=0.12cm
    \caption{USC-GRAD-STDdb results}
    \label{tab:resultSTD}
    \centering
    \begin{tabular}{l|ccccccc}
    \hline
    ~& \rot{Drone} & \rot{Boat} & \rot{Vehicle} & \rot{Person} & \rot{Bird} & \specialcell{Mean\\AP$_{0.5}$} &\specialcell{Mean\\AP$_{0.5-0.95}$}\\ \hline
    \specialcellSTD{FPN-X101\\\quad{\scriptsize baseline}} & 65.2& 44.9 & 36.1 & 62.8 & 13.4 & 44.5 & 16.3 \\
    \specialcellSTD{FANet\\\quad{\scriptsize(Short-Term)}} & 67.9 & 49.5 & 38.9 & 66.2 & 20.3 & 48.5 & 17.6 \\
    \specialcellSTD{FANet\\\quad{\scriptsize (Short\&Long-Term)}}& 66.8 & 53.3 & 35.9 & 68.4 & 25.1 & 49.9 & 18.3\\ \hline
    \end{tabular}
\end{table}

We show the resuts on the USC-GRAD-STDdb dataset in Table~\ref{tab:resultSTD}. Our method outperforms the single frame baseline in every object category just using short-term information. Unlike \cite{bosquet2018stdnet} and \cite{bosquet2020stdnet} that give a class agnostic mAP, we report the results taking into account the object categories. Adding both long-term and short term-information leads to 49.9\% mAP with an IoU threshold of 0.5. Our system improves the single frame baseline (FPN-X101 with cascade head) by 5.4\% in this metric. We also compare the average mAP changing the IoU threshold from 0.5 to 0.95 with a step of 0.05, improving the single frame baseline by 2.0\%.

In general, working with objects smaller than $16\times16$ pixels, represents a challenge for our spatio-temporal approach due to the relative movement of the objects with respect to their size: a small displacement of a few pixels might cause the object overlap among neighbouring frames to decrease dramatically, making harder to build accurate tubelet proposals from \emph{anchor cuboids}. Therefore, we reduce the tubelet length, working with N=4 for all experiments on the USC-GRAD-STDdb dataset. In this line, the most challenging class of the dataset is \emph{Bird}, as objects in this category are generally fast moving and with highly irregular trajectories. Nevertheless, analyzing the per category results, the improvement in the \emph{Bird} class boosting the AP from 13.4\% to 25.1\% is very promising and shows the benefits of adding spatio-temporal information to the small object detection task.

\section{Conclusion}
We have described a novel framework that extends two stage object detectors to exploit spatio-temporal information to improve object detection in videos. We have proposed a feature aggregation method throughout short tubelet proposals without any other external information, such as associations given by tracking algorithms or optical flow. We have redesigned the network head, to take advantage of this aggregated spatio-temporal data, with a double head implementation specialized in both spatial and spatio-temporal information. Our experimentation has proved that this short-term information is complementary to the long-term information calculated by our linking method. The overall framework achieves competitive results in the widely used ImageNet VID dataset with 80.9\% mAP. Also, it significantly outperforms the single frame baseline in the USC-GRAD-STDdb dataset.

Although our tubelet initialization based on anchor cuboids provides a light computational method to link objects through neighbour frames, it imposes a limitation in the maximum tubelet length. In the future, we will further develop this component making it more flexible in order to avoid this limitation.

\section*{Acknowledgment}
This research was partially funded by the Spanish Ministry of Science, Innovation and Universities under grants TIN2017-84796-C2-1-R and RTI2018-097088-B-C32, and the Galician Ministry of Education, Culture and Universities under grants ED431C 2018/29, ED431C 2017/69 and accreditation 2016-2019, ED431G/08. These grants are co-funded by the European Regional Development Fund (ERDF/FEDER program).

\ifCLASSOPTIONcaptionsoff
  \newpage
\fi

\bibliographystyle{IEEEtran}
\bibliography{IEEEabrv, references}

\end{document}